\documentclass[letterpaper,10pt,journal]{IEEEtran}
\IEEEoverridecommandlockouts 

\usepackage{xparse}
\usepackage{color}
\usepackage{booktabs}
\usepackage{multirow}
\usepackage{multicol}
\usepackage{graphicx} 
\usepackage{comment}
\usepackage{hyperref}
\usepackage{array}
\usepackage{subcaption}
\usepackage{caption}
\usepackage{flushend}
\graphicspath{{figs/}}

\usepackage{amsmath}
\usepackage{amssymb}
\usepackage{amsfonts}
\usepackage{mathrsfs}
\usepackage{mathtools}
\usepackage{bm}
\allowdisplaybreaks

\usepackage{balance}
\usepackage{cleveref}
\crefformat{equation}{(#2#1#3)}

\Crefname{figure}{Fig.}{Figs.}

\newcommand{\ts}{\hspace*{0.3em}}

\usepackage[backend=biber,hyperref=true,style=ieee-caps]{biblatex}
\usepackage{xparse}

\NewDocumentCommand\bbm{}{ \begin{bmatrix} }
\NewDocumentCommand\ebm{}{ \end{bmatrix} }
\NewDocumentCommand\Vector{m}{ \boldsymbol{\mathbf{#1}} }

\NewDocumentCommand\Matrix{m}{ \bm{\mathbf{#1}} }
\NewDocumentCommand\Transpose{m}{ \left.{#1}\right.^{\mathsf{T}} }

\NewDocumentCommand\Real{}{ \mathbb{R} }

\NewDocumentCommand\Wedge{m}{\left(#1\right)^\wedge}
\NewDocumentCommand\Vee{m}{\left(#1\right)^\vee}

\NewDocumentCommand\LieGroupSO{m}{ \mathrm{SO}(#1) }
\NewDocumentCommand\LieAlgebraSO{m}{ \mathfrak{so}(#1) }

\NewDocumentCommand\LieGroupSE{m}{ \mathrm{SE}(#1) }

\NewDocumentCommand\Matexp{m}{\mathrm{exp}\left({#1}\right)}

\NewDocumentCommand\Expectation{m}{ \mathbb{E}\left[#1\right] }
\NewDocumentCommand\NormalDistribution{mm}{ \mathcal{N}\left(#1,#2\right) }

\NewDocumentCommand\Zero{}{ \Matrix{0} }

\NewDocumentCommand\Mean{m}{\bar{#1}}

\NewDocumentCommand\Inv{m}{#1^{-1}}
\NewDocumentCommand\Defined{}{\triangleq}

\NewDocumentCommand{\LieGroupGL}{m}{ \mathrm{GL}\left(#1\right) }

\NewDocumentCommand{\LieGroupGal}{m}{ \mathrm{Gal}\left(#1\right) }

\NewDocumentCommand{\LieGroupSGal}{m}{ \mathrm{SGal}\left(#1\right) }
\NewDocumentCommand{\LieAlgebraSGal}{m}{ \mathfrak{sgal}\left(#1\right) }
\NewDocumentCommand{\LieGroupSETwo}{m}{ \mathrm{SE_{2}}\left(#1\right) }

\let\originalleft\left
\let\originalright\right
\renewcommand{\left}{\mathopen{}\mathclose\bgroup\originalleft}
\renewcommand{\right}{\aftergroup\egroup\originalright}

\NewDocumentCommand{\<}{}{\mspace{1mu}}

\makeatletter
\newcommand{\vast}{\bBigg@{3}}
\newcommand{\Vast}{\bBigg@{4}}
\makeatother

\title{\huge Making Space for Time:\ts The Special Galilean Group\\ and Its Application to Some Robotics Problems}

\author{Jonathan Kelly$^\dagger$\vspace*{-8mm} 
\thanks{$^\dagger$\hspace{0.2em}Jonathan Kelly is with the Space \& Terrestrial Autonomous Robotic\linebreak Systems (STARS) Laboratory, University of Toronto Institute for Aerospace Studies (UTIAS), Toronto, Canada. {\tt\footnotesize jkelly@utias.utoronto.ca}}}

\begin{document}
\maketitle

\begin{abstract}
The special Galilean group, usually denoted SGal(3), is a 10-dimensional Lie group whose important subgroups include the special orthogonal group, SO(3), the special Euclidean group, SE(3), and the group of extended poses, SE$_{\text{2}}$(3).
We briefly describe SGal(3) and its Lie algebra and show how the group structure supports a unified representation of uncertainty in space and time.
Our aim is to highlight the potential usefulness of this group for several robotics problems.
\end{abstract}

\section{Introduction}
\label{sec:intro}

Inertial reference frames play a central role in mechanics and, by extension, in robotics.
An inertial frame is (roughly) a non-accelerating, non-rotating frame; any frame moving at a constant velocity relative to an inertial frame is also inertial.
Relationships between inertial frames, and between events (i.e., points in space and time) observed in different inertial frames, may be described by Galilean transformations, which include translations in space and time, rotations of spatial coordinates, and Galilean velocity boosts \cite{2011_Holm_Geometric_Part_II}.

The set of Galilean transformations forms a 10-dimensional Lie group, called the Galilean group and denoted $\LieGroupGal{3}$, that is the symmetry group of Galilean relativity.
We consider the special Galilean group, $\LieGroupSGal{3}$, which is the connected component of $\LieGroupGal{3}$ at the identity element.
Notably, many of the important groups applied frequently to robotics problems are proper subgroups of the special Galilean group, including the special orthogonal group, $\LieGroupSO{3}$, the special Euclidean group, $\LieGroupSE{3}$, and the group of extended poses, $\LieGroupSETwo{3}$ \cite{2020_Barrau_Mathematical}.
However, $\LieGroupSGal{3}$ is distinct in that time is one dimension of the group manifold.

Perhaps surprisingly, despite being well known in the geometric mechanics literature (e.g., treated in \cite{1999_Marsden_Introduction} and in other popular texts), the special Galilean group has received little attention in robotics.
Our intent in this short paper is to introduce $\LieGroupSGal{3}$ and its Lie algebra $\LieAlgebraSGal{3}$ and to provide a few suggestions for possible applications to some robotics problems. 
We focus on the relationship between temporal and spatial uncertainty.
Specifically, uncertainty in space is related to uncertainty in time through motion; said differently, uncertainty about \emph{when} leads to uncertainty about \emph{where}.

\begin{figure*}[t]
\centering
\vspace{-4.75mm}
\includegraphics[height=1.7in]{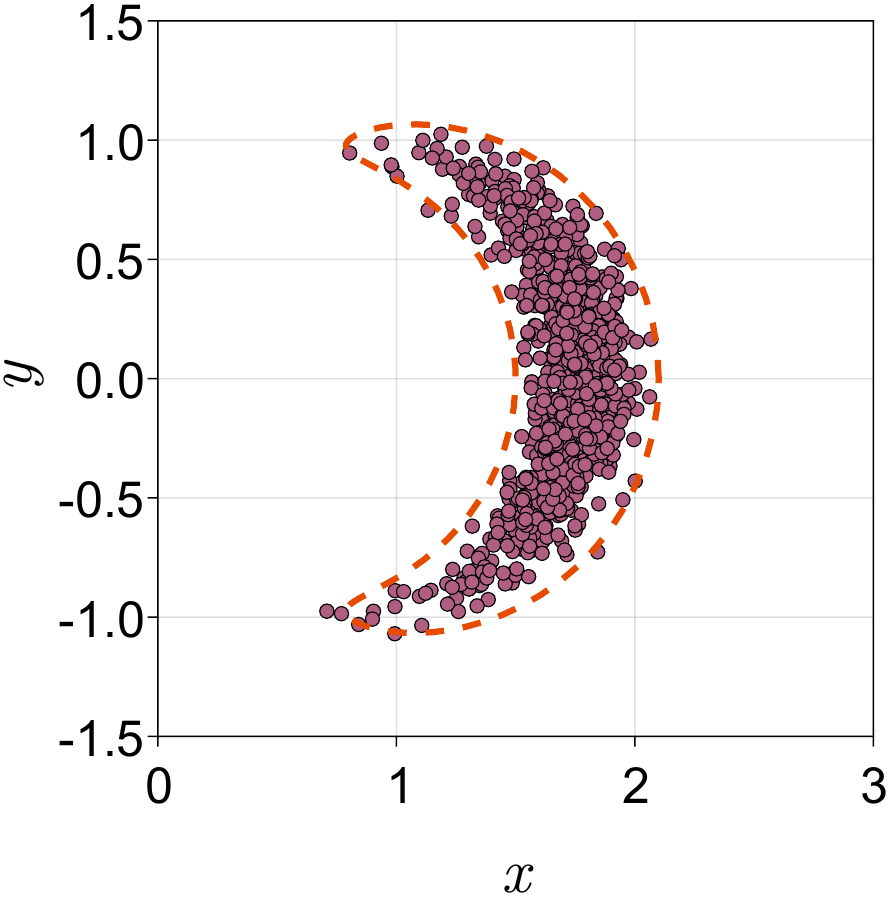}
\hspace{12mm}
\includegraphics[height=1.7in]{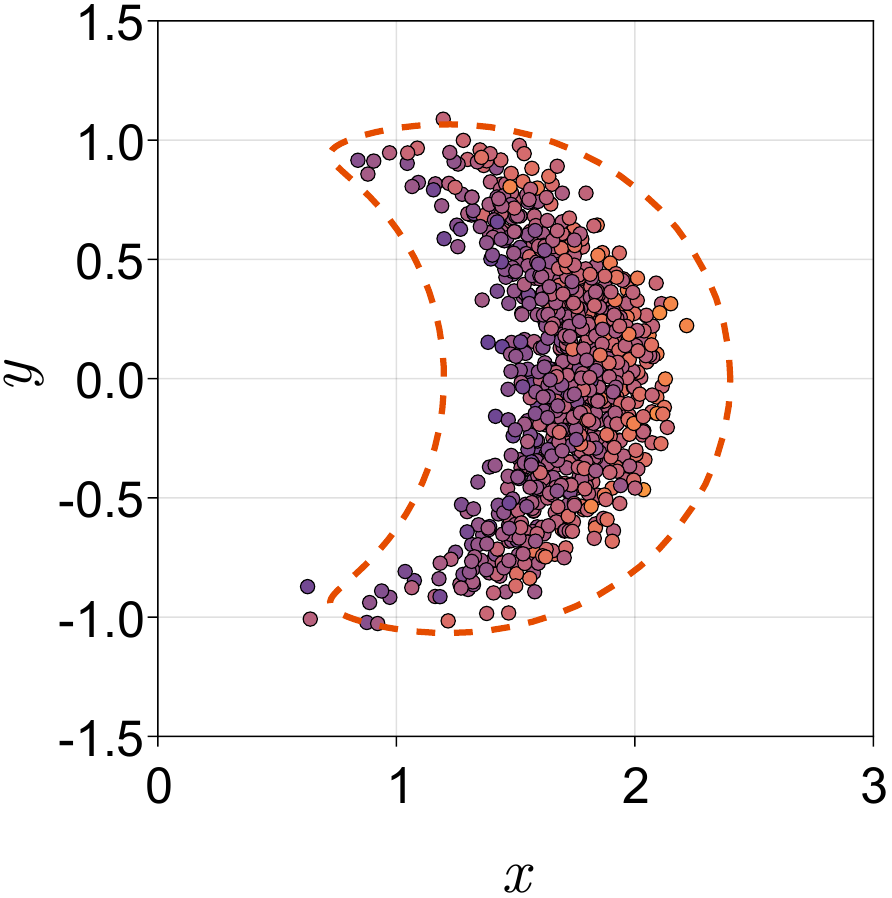}
\hspace{12mm}
\includegraphics[height=1.7in]{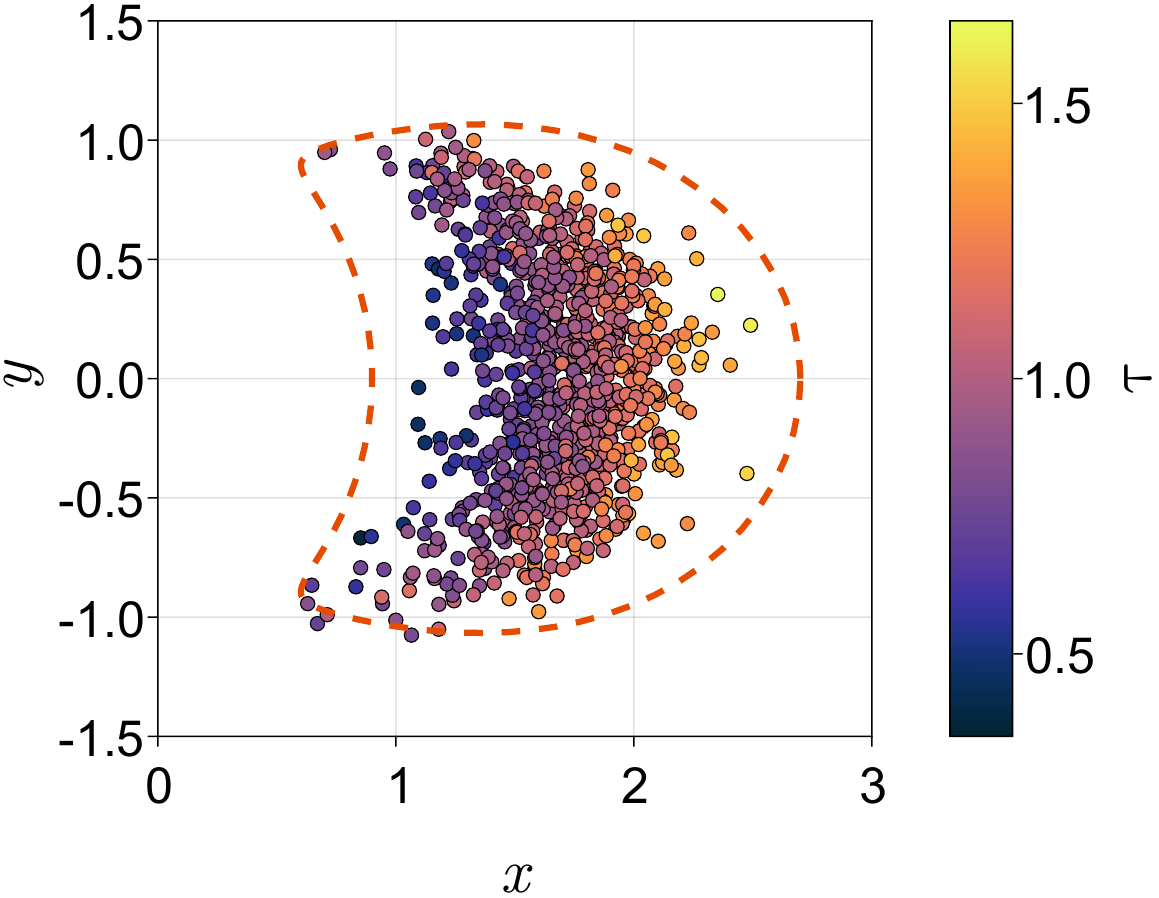}
\hspace{1mm}
\caption{Visualization of the transformation of (the coordinates of) an event by a randomly right-perturbed element of $\LieGroupSGal{3}$, projected onto the $x$-$y$ plane. The moving reference frame has a nonzero velocity in the positive $x$ direction. \textit{Left:} Perturbation to $x$ translation and $z$ orientation components only. \textit{Middle:} Added (small) perturbation in time. \textit{Right:} Added (larger) perturbation in time. Each plot shows 1,000 samples drawn from a multivariate Gaussian, shaded by temporal offset. The dashed red line is the projection onto the $x$-$y$ plane of the 3$\sigma$ bounds of the transformed Gaussian distribution.}
\label{fig:uncertainty}
\vspace{-4mm}
\end{figure*}

\section{The Matrix Lie Group\hspace*{0.05em} $\LieGroupSGal{3}$}
\label{sec:group}

Galilean relativity unifies space and time into a single, four-dimensional manifold, where points on the manifold, called \emph{events}, are specified by three spatial coordinates and one temporal coordinate, $\left(\Vector{x}, t \right) \in \Real^{3} \times \Real$.
An event can also be specified as a five-element homogeneous column, $\Vector{p} = \smash{\bbm \Transpose{\Vector{x}} &\!\! t &\!\! 1 \ebm^{\mathsf{T}}}$.
The (left) action of \(\LieGroupSGal{3}\) on the set of events \(\mathcal{E}\) is a map \(\alpha : \LieGroupSGal{3} \times \mathcal{E} \rightarrow \mathcal{E}\) that preserves absolute time intervals and spatial distances between simultaneous events, reflecting the symmetries of Galilean relativity.

Elements of $\LieGroupSGal{3}$ can be written as 5$\times$5 matrices,
\begin{equation}
\label{eqn:SGal3_definition}
\Matrix{F} =
\bbm
\Matrix{C} & \Vector{v} & \Vector{r} \\ 
\Zero & 1 & \tau \\
\Zero & 0 & 1
\ebm  
\in \Real^{5 \times 5},
\end{equation}
where $\Matrix{C} \in \LieGroupSO{3}$ is a rotation matrix, $\Vector{v} \in \Real^3$ is a velocity vector, $\Vector{r} \in \Real^3$ is a position vector, and $\tau \in \Real$ is a time.
The matrix form is an inclusion $\LieGroupSGal{3} \rightarrow \LieGroupGL{5, \Real}$.
The group has the semidirect product structure $\LieGroupSGal{3} = \left(\LieGroupSO{3} \ltimes \Real^{3}\right) \ltimes \left(\Real^{3} \times \Real \right)$.
An element $\Matrix{F} \in \LieGroupSGal{3}$ acts on an event $\Vector{p}$ (expressed as a column) such that $\Vector{p}' = \Matrix{F}\Vector{p}$, transforming from the local frame (on the right) to the global frame (on the left).

The set of all tangent vectors at the identity of $\LieGroupSGal{3}$ defines the Lie algebra $\LieAlgebraSGal{3}$.
This tangent space is a 10-dimensional vector space whose elements can also be written as 5$\times$5 matrices,
\begin{equation}
\label{eqn:sgal3_definition}
\Matrix{\Xi} = \hspace{1pt}
\bbm
\Vector{\phi}^{\wedge} & \Vector{\nu} & \Vector{\rho} \\
\Zero & 0 & \iota \\
\Zero & 0 & 0
\ebm
\in \Real^{5 \times 5},
\end{equation}
where $\Vector{\phi}^{\wedge} \in \LieAlgebraSO{3}$ is a skew-symmetric matrix, $\Vector{\nu} \in \Real^3$, $\Vector{\rho} \in \Real^3$, and $\iota \in \Real$.
An element of $\LieAlgebraSGal{3}$ can be considered as the `velocity' along a curve $\gamma: \Real \rightarrow \LieGroupSGal{3}$ passing through the identity at $\gamma(0)$.\footnote{Note that, in general, $\gamma$ is \emph{not} parameterized by time (because time is part of the $ \LieGroupSGal{3}$ manifold). In turn, an arbitrary vector $\Vector{\nu}$, for example, should not be treated as defining `acceleration with respect to time.'}
The linear `wedge' operator $\Wedge{\cdot}$ maps $\Real^{10} \rightarrow \LieAlgebraSGal{3}$,
\begin{equation}
\Vector{\xi}^{\wedge}
=
\bbm
\Vector{\rho} \\
\Vector{\nu}  \\
\Vector{\phi} \\
\iota
\ebm^{\wedge}
=
\bbm
\Vector{\phi}^{\wedge} & \Vector{\nu} & \Vector{\rho} \\
\Zero & 0 & \iota \\
\Zero & 0 & 0
\ebm.
\end{equation}
The exponential map $\exp : \LieAlgebraSGal{3} \rightarrow \LieGroupSGal{3}$ is
\begin{equation}
\label{eqn:SGal3_exp_map_short}
\vspace*{0.5mm}
\exp\left(\Vector{\xi}^{\wedge}\right)
=
\sum_{n = 0}^{\infty}\frac{1}{n!}
\left(\Vector{\xi}^{\wedge}\right)^{n}
=
\bbm
\Matrix{C} & 
\Matrix{D}\<\Vector{\nu} & 
\Matrix{D}\Vector{\rho} +
\Matrix{E}\<\Vector{\nu}\iota \\
\Zero & 1 & \iota \\
\Zero & 0 & 1
\ebm,
\vspace*{0.5mm}
\end{equation}
where the matrices $\Matrix{C}$, $\Matrix{D}$, and $\Matrix{E}$ are all available in closed form.
Importantly, the term $\Matrix{E}\<\Vector{\nu}\iota$ that is part of the upper-right entry of the matrix in \Cref{eqn:SGal3_exp_map_short} does not appear in the exponential map for $\LieGroupSETwo{3}$.
Complete derivations of the exponential and logarithmic maps (including the submatrices appearing in \Cref{eqn:SGal3_exp_map_short}), the adjoint representations of $\LieGroupSGal{3}$ and $\LieAlgebraSGal{3}$, and the Jacobian of $\LieGroupSGal{3}$, are provided in \cite{2023_Kelly_Galilean}.


\section{Uncertainty on $\LieGroupSGal{3}$}
\label{sec:uncertainty}

The uncertainty associated with an element of $\LieGroupSGal{3}$ can be expressed in terms of a perturbation in the tangent space. 
Following the approach in \cite{2014_Barfoot_Associating}, we assume that the perturbation is a vector-valued Gaussian random variable, $\Vector{\xi} \sim \NormalDistribution{\Vector{0}}{\Matrix{\Sigma}}$. The perturbation can be applied locally (on the right) or globally (on the left),
\begin{equation}
\Matrix{F} = \Mean{\Matrix{F}}\,\Matexp{\Vector{\xi}^{\wedge}}
\quad\text{or\hspace*{0.2em}}\quad
\Matrix{F} = \Matexp{\Vector{\xi}^{\wedge}}\,\Mean{\Matrix{F}},
\end{equation}
respectively, where $\Mean{\Matrix{F}}$ is the mean of the distribution $p\left(\Matrix{F}\right)$ induced over $\Matrix{F}$ by $\Vector{\xi}$. If we consider a local perturbation, we can write the covariance $\Matrix{\Sigma}$ of the Gaussian as the expectation
\begin{equation}
\label{eqn:covariance}
\Matrix{\Sigma}
\Defined
\Expectation{\Vector{\xi}\Transpose{\Vector{\xi}}} 
= 
\Expectation{
\ln\left(\Inv{\Mean{\Matrix{F}}}\Matrix{F}\right)^{\vee} %
\Transpose{\ln\left(\Inv{\Mean{\Matrix{F}}}\Matrix{F}\right)^{\vee}}}
\in \Real^{10 \times 10},
\end{equation}
where $\mathrm{ln} : \LieGroupSGal{3} \rightarrow \LieAlgebraSGal{3}$ is the logarithmic map (see \cite{2023_Kelly_Galilean} for details).
The `vee' operator $\Vee{\cdot}$ maps $\LieAlgebraSGal{3} \rightarrow \Real^{10} $.

\Cref{fig:uncertainty} visualizes how an event, specified in the local frame, is transformed by a randomly-perturbed element of $\LieGroupSGal{3}$.
The left plot depicts the case where only the $x$ translation and $z$ orientation components of the transformation are perturbed, yielding the well-known `banana-shaped' distribution on $\LieGroupSE{3}$ \cite{2011_Chirikjian_Stochastic,2012_Long_Banana}.
The middle and right plots show cases where additional (small and large, respectively) random perturbations are applied to $\tau$; temporal uncertainty induces a spread in the spatial uncertainty that is velocity-dependent.

\section{Applications}
\label{sec:apps}

The group $\LieGroupSGal{3}$ is a natural fit for problems involving the propagation of motion estimates over time and for situations in which temporal uncertainty is important.
We very briefly review three applications where $\LieGroupSGal{3}$ is relevant: preintegration, navigation, and calibration.
To the best of our knowledge, the only (robotics-specific) work to date on these topics is \cite{2019_Fourmy_Absolute} and \cite{2021_Giefer_Uncertainties}.

\subsection{Preintegration}
\label{subsec:preintegration}

High-rate inertial (IMU) data can be integrated over time to yield an incremental pose change (i.e., a pose delta).
Preintegration schemes on $\LieGroupSO{3} \times \Real^{3}$ \cite{2017_Forster_On-Manifold} and on $\LieGroupSETwo{3}$ \cite{2022_Brossard_Associating} have appeared in the literature.
However, preintegration is more accurately (and elegantly) formulated on $\LieGroupSGal{3}$, as shown by Fourmy et al.\ in \cite{2019_Fourmy_Absolute}.
If the IMU angular rates and linear accelerations are treated as constant over a short time interval $\Delta t$, then the pose change is obtained directly as
\begin{equation}
\label{eqn:imu_delta}
\Matrix{F}_{\Delta t}
=
\Matexp{
\left(
\Transpose{\bbm
\Transpose{\Zero} &
\Transpose{\Vector{a}} &
\Transpose{\Vector{\omega}} &
1
\ebm}
\Delta t
\right)^{\wedge}
},
\end{equation}
where $\Vector{\omega}$ and $\Vector{a}$ are the angular rates and linear accelerations, respectively, and $\Matrix{F}_{\Delta t} \in \LieGroupSGal{3}$.\footnote{We neglect the IMU biases here for the sake of brevity, but they are straightforward to incorporate (see \cite{2019_Fourmy_Absolute} for details).}
While the approach in \cite{2019_Fourmy_Absolute} does not explicitly consider uncertainty in time, it still outperforms preintegration on $\LieGroupSETwo{3}$, for example, because \Cref{eqn:imu_delta} properly captures the geometry of the problem.


\subsection{Navigation and Calibration}
\label{subsec:motion}

Most motion estimation algorithms consider the uncertainty of spatial measurements (e.g., position or orientation) but treat measurement timestamps as exact (without uncertainty).
However, measurement times may also be subject to non-negligible errors and to random noise.
By accounting for temporal uncertainty, it should be possible to increase the accuracy of motion estimates.
Some preliminary work in this direction is presented by Giefer in \cite{2021_Giefer_Uncertainties}, but for $\LieGroupSGal{2}$ only.
One important consideration is that errors in time may not be Gaussian-distributed, necessitating the use of more complex estimation methods.

Given that most modern robots are equipped with multiple sensors, accurate multisensor calibration is essential for reliable operation. 
Notably, unknown delays between sensor data streams can adversely affect state estimates. 
The structure of $\LieGroupSGal{3}$ appropriately describes the (coupled) geometry of spatiotemporal calibration, when any delays (and measurement timestamps) are uncertain.
Our own prior work on camera-IMU calibration handled temporal and spatial uncertainty separately \cite{2014_Kelly_General,2014_Kelly_Determining}; we believe there is potential for improvement by developing a unified approach based on the use of $\LieGroupSGal{3}$.


\section*{Acknowledgements}
\label{sec:ack}

The author would like to thank Hugh Corley for suggesting the title of this paper.
This work was supported in part by the Canada Research Chairs Program.

\printbibliography

\end{document}